\documentclass[runningheads]{llncs}
\usepackage[T1]{fontenc}
\usepackage{graphicx}
\PassOptionsToPackage{table}{xcolor} 
\usepackage{amssymb,amsmath}
\usepackage[labelfont=bf]{caption}
\usepackage{subcaption}
\usepackage{xcolor}
\usepackage{soul}
\usepackage{multirow}
\usepackage{multicol}
\usepackage{booktabs}
\usepackage{cite}
\usepackage{mathrsfs}
\usepackage{listings}
\usepackage{parskip}
\usepackage[shortlabels]{enumitem}
\usepackage{tcolorbox}
\usepackage{dsfont}
\usepackage{algorithm}
\usepackage{algpseudocode}
\usepackage{fancyvrb}
\usepackage{array}
\usepackage{makecell}

\graphicspath{{images/}}

\usepackage[breaklinks, pagebackref, colorlinks]{hyperref}
\usepackage{color}

\begin{document}
\title{Quality Control for Radiology Report Generation Models via Auxiliary Auditing Components}

\titlerunning{Error Detection for Report Generation Models}
%
\author{Hermione Warr\inst{1}, 
Yasin Ibrahim\inst{1}, 
Daniel R. McGowan\inst{2,3}, \\ 
Konstantinos Kamnitsas\inst{1,4,5}}
\authorrunning{H. Warr et al.}
\institute{Department of Engineering Science, University of Oxford, Oxford, UK \\
\email{\{first\_name.last\_name\}@eng.ox.ac.uk}
\and
Department of Oncology, University of Oxford, UK
\and
Department of Medical Physics and Clinical Engineering, Oxford University Hospitals NHS FT, Oxford, UK
\and
Department of Computing, Imperial College London, London, UK
\and
School of Computer Science, University of Birmingham, Birmingham, UK
}

%

\maketitle              
\begin{abstract}
Automation of medical image interpretation could alleviate bottlenecks in diagnostic workflows, and has become of particular interest in recent years due to advancements in natural language processing. 
Great strides have been made towards automated radiology report generation via AI, yet ensuring clinical accuracy in generated reports is a significant challenge, hindering deployment of such methods in clinical practice. 
In this work we propose a quality control framework for assessing the reliability of AI-generated radiology reports with respect to semantics of diagnostic importance using modular auxiliary auditing components (ACs).
Evaluating our pipeline on the MIMIC-CXR dataset, our findings show that incorporating ACs in the form of disease-classifiers can enable auditing that identifies more reliable reports, resulting in higher F1 scores compared to unfiltered generated reports. 
Additionally, leveraging the confidence of the AC labels further improves the audit's effectiveness.
Code will be made available at: \url{https://github.com/hermionewarr/GenX_Report_Audit}

\end{abstract}
\section{Introduction}
\label{sec:intro} 
Radiologists are facing mounting challenges managing the growing volume of medical imaging data under resource constraints, necessitating more efficient and innovative solutions for image analysis and interpretation. Automated (or partly automated) radiology report generation is one such solution.
Since the introduction of transformers~\cite{vaswani_attention_2017}, which significantly enhanced language generation, numerous studies have investigated radiology report generation from medical images, particularly chest X-rays (CXR)~\cite{chen-2020, Hou_ratchet_2021, Serra_2023, llava-med_2023, moor_med-flamingo_2023, tanida_interactive}.
However, deployment of language models to medical applications faces a variety of challenges, with a key area being the need for factual accuracy. 
Various methods have been developed to address this - particularly useful have been the introduction of clinical metrics to supplement the standard natural language processing (NLP) metrics used to evaluate report generation quality \cite{chen-2020, liu_2019}.
Previous efforts to improve the quality of the generated reports have incorporated extra information during training, such as bounding boxes for improved localisation, or have aimed to refine diagnostic accuracy by including loss functions that reward clinical accuracy \cite{tanida_interactive, Chen_xmod_2021, miura_fact_comp_2021, Serra_2023}.
Additionally, more complex architectures, such as memory-augmented attention, have been developed to facilitate better image understanding \cite{miura_fact_comp_2021, meshed-memory_T_2020}. 

Besides efforts to improve report generation accuracy, another line of work is error detection in language modeling. Most work in this area has targeted the natural language domain. The two main types of uncertainty estimation in large language models (LLMs) are statistical uncertainty (SU) and in-dialogue uncertainty (IDU) \cite{Tomani_fb_unc_2024}. The latter relies on large pretrained models to admit when they don't know an answer \cite{Li_conf_llm_2024, Kostumov_uncer_vlp_2024}.
SU relates certainty to the model's output entropy, though in language modelling, it must address the coherence of entire sentences, as well as on a per token basis \cite{Malinin_Gales_2021,Kuhn_OATML_2023}. These methods often depend on extensive labelled datasets in the natural language domain, which do not adequately capture the specialised terminology and context of clinical settings \cite{Tomani_fb_unc_2024}. 
Few works have investigated the possibility of error detection in reports written by radiologists \cite{Zech_2019, Wu_ucl_er_2024, Kim_ucl_eye_2024, gpt_4_rad_er}. 
To our knowledge, no previous work has investigated error detection for \emph{automated} report generation, which is the focus of our study.

We introduce an auditing framework for identifying potential errors in AI-based radiology report generation, using modular auxiliary auditing components (AC) for quality control. Our study focuses on chest-Xray (CXR) imaging, defining ACs as image-based disease classifiers to extract semantics relevant to diagnosis. We develop a report generation model, GenX, which performs competitively with previous works and serves as basis for testing our auditing framework. Reports are audited using ACs that provide disease classification and confidence levels. Reports with consistent diagnoses across the pipeline meet our criteria, while inconsistencies flag potential errors. Additionally, by leveraging the ACs' classification confidence, we enforce stricter quality control, requiring consistency with the highest confidence predictions. Our evaluation on MIMIC-CXR shows a significantly higher F1 score for reports meeting the audit's criteria. Results demonstrate that consistency between report and auxiliary components is a promising auditing mechanism for automated report generation.

\section{Methodology}

\begin{figure}
\caption{Proposed error detection pipeline of radiology report generation using auxiliary auditing components. The standard report generation pipeline ($a$) is followed by the CheXbert labeler, $g_T$ ($b$) that extracts pathology labels, $C_T$, from the reports, that are semantically meaningful for clinical diagnosis. ($c$) Modular image-based audit models, AC, here disease classifiers, predict disease class labels, $C_I$, based on image, $I$. ($d$) If the labels predicted based on image ($C_I$) and report ($C_T$) are consistent, the report's contents are deemed likely reliable. In case of inconsistency, $C_I\neq C_T$, 
the report is flagged as less reliable, potentially containing an error. If the image-based classifiers have predictive confidence below threshold $t$, auditing can be deferred to user due to uncertainty.}
    \centering
    \includegraphics[width=\textwidth]{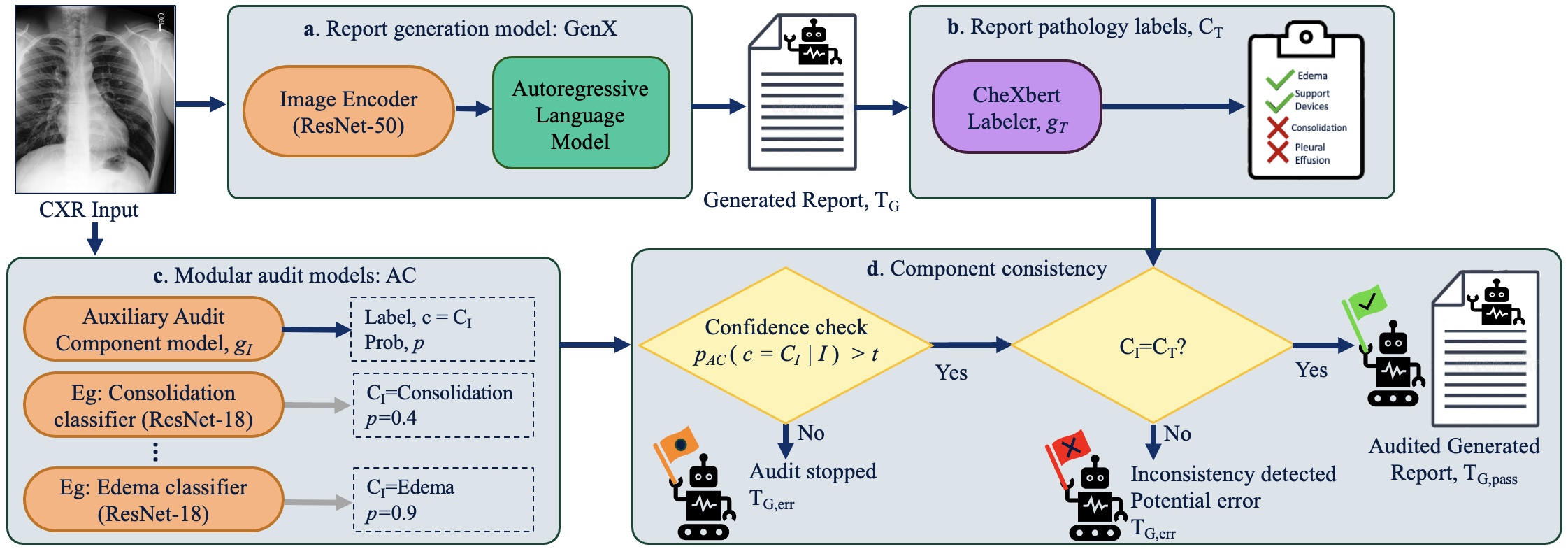}
    \label{fig:pipeline}
\end{figure}

\subsection{Report Generation Model - GenX}
The proposed framework for auditing radiology report generation is shown in Fig.~\ref{fig:pipeline}. The backbone of the framework is the report generation model. The generalised vision language modeling problem of generating a textual sequence T given an input image I can be framed as estimating the conditional distribution $p(T|I)$, as a product of the conditional probabilities:
\begin{equation}
p(T|I) = \prod_{l=1}^{L} p(T_l\:|\:T_{<l},\:I;\:\theta),
\label{eq:prob}
\end{equation}
where $T$ is modelled as a sequence of word tokens $\{T_{1},...,T_{L}\}$, with $T_{<l}$ being the set of tokens preceding $T_l$, where $L$ is the number of tokens in the sequence, and $\theta$ the model parameters \cite{moor_med-flamingo_2023, 2022_Flamingo}.

\begin{figure}
\caption{Illustration of one inference step for radiology report generation (a). 
Embeddings of image and previously generated text tokens are passed to the autoregressive language model ($b$) which outputs a probability distribution over the vocabulary, to predict the next word in the sequence.}
\centering
\includegraphics[width=\textwidth]{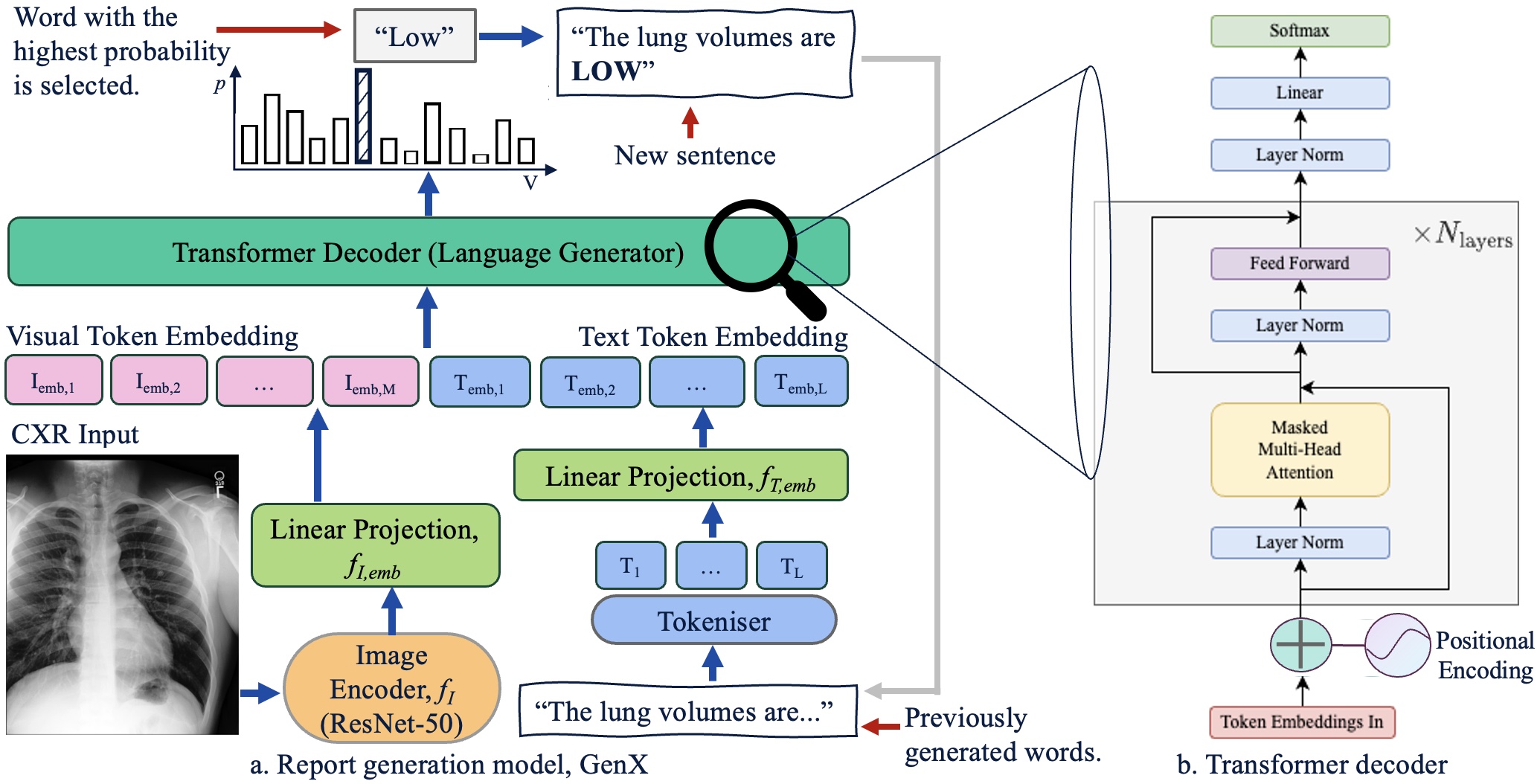}%
\label{fig:model}
\end{figure}

An overview of our report generation model, \textit{GenX}, is shown in Fig.~\ref{fig:model}. The model is trained to maximise the probability of generating the target sequence from the image, $p(T|I)$ from eq~\ref{eq:prob}.
Image features $I_{feat}=f_I(I) \in \mathbb{R}^{D_I}$, with dimensionality $D_I$, are extracted by image encoder $f_I$.
A linear layer, $f_{I,emb}$, projects $I_{feat}$ to the space of token embeddings that will be input to the language generator, giving $I_{emb}=\{I_{emb,1} \ ...\ I_{emb,M}\}=f_{I,emb}(I_{feat}) \in \mathbb{R}^{D_T \times M}$, a set of $M$ image token embeddings with dimensionality $D_T$, similarly to LLaVA~\cite{llava_2023}. 
A tokeniser maps text to a set of $L$ token vectors $\{T_l\}_{1:L} \in \mathbb{R}^{L \times V}$, where $V$ is the vocabulary size. Each $T_l$ is embedded by a linear layer to $T_{l,emb}=f_{T,emb}(T_l) \in \mathbb{R}^{D_T}$.
The concatenated embeddings of image and text tokens $\{I_{emb,1}\ ...\ I_{emb,M}, T_{emb,1}\ ...\ T_{emb,L}\}$ are passed to a Transformer decoder (Fig.~\ref{fig:model}b) which predicts a probability distribution over $V$ vocabulary tokens, expressing which is more probable in the sequence. The decoder has $N_{layers}$ with Attention blocks that each has $N_{heads}$ parallel heads \cite{vaswani_attention_2017}.
The model is trained using the common paradigm of diagonal attention masking \cite{vaswani_attention_2017}.
A report $T_G$ is generated with auto-regressive inference, starting only with image token embeddings $I_{emb}$ as input, then predicting the first and each following word token as the one with highest predicted probability.

\subsection{Auditing Generated Reports via Auxiliary Classifiers}
\label{subsec:method_audit}

While some types of mistakes in AI-generated radiology reports, such as grammatical errors, are merely inconvenient, semantic mistakes related to disease diagnosis can be detrimentalor patient care. 
Consider a semantic concept of interest, denoted $c$. We focus on the case where $c$ represents a specific disease, with the class label $C\!\in\!\{1,0\}$ indicating the presence or absence of that disease. 
To audit whether a generated report $T_G$ is semantically correct about $c$, 
we need to extract the value $C_T$ described in text $T_G$ via a mapping function $g_{T}\!:\! T_G \!\rightarrow\! c$. This can be any text-based classifier, or prompt-based LLM that can answer questions in the form "Does this text suggest evidence or not for disease $c$?", which have become widely available. Herein we use a pretrained CheXbert text-classifier \cite{chexbert_2020} as $g_{T}$, obtaining $C_T = g_{T}(T_G)$ (Fig~\ref{fig:pipeline}b).

We then introduce to the framework an independent model $g_I\!:\! I \!\rightarrow\!c$, which infers value $C_I$ for concept of interest $c$ from image $I$ (Fig.~\ref{fig:pipeline}c). We term such models auxiliary auditing components (AC). In the examined case when $c$ is existence of specific disease, $g_I$ is a classifier predicting $C_I\!=\!1$ if image $I$ shows the disease or $C_I\!=\!0$ otherwise, with model confidence $p_{AC}(c\!=\!C_I |I)$.
If multiple concepts $c$ are of interest (e.g. different diseases), an independent AC model can be used per $c$, for ease of modular development, or a multi-label AC model.

Auditing is performed by comparing values $C_T$ and $C_I$ about the concept $c$ of interest (e.g. disease) extracted from the report or the image respectively (Fig~\ref{fig:pipeline}d). To consider the contents of the report reliable with respect to $c$, agreement between the report and AC is required, i.e. $C_T\! =\! C_I$. We further extend this by taking into consideration the confidence $p_{AC}$ of the AC. We only consider the auditing successful if it satisfies the additional requirement that the value $C_T=C_I$ has been inferred by the AC with confidence $p_{AC}$ over a pre-determined threshold $t$. Formally, for a successfully completed auditing we require: 
\begin{equation}
    (C_I = C_T) \land (p_{AC}(c = C_I\ |\ I) \geq t), 
\label{eq:condition}
\end{equation}
where $\land$ is the logical \emph{and} operator. 
The set of generated reports that pass the audit for a particular concept $c$ is denoted by $T_{G,pass}^{(c)}=\{T_G | \ \text{Eq.~\ref{eq:condition} is } True\}$. The remainder, $T_{G,err}^{(c)} \!=\! \{T_G | \ \text{Eq.~\ref{eq:condition} is } False\}$, can be deferred to the human user for review as potentially inaccurate for that disease.

The framework is designed based on ACs that are modular, independent models. The design draws inspiration from N-Version programming \cite{chen1978n} and the principle of redundancy for reliability in software engineering. Assuming independence of components ($GenX$, $g_T$, $g_I$) with label error rates $e_{C_T}$ and $e_{C_I}$, the error rate of the auditing framework 
is $e=e_{C_T} \cdot e_{C_I} $, which is lower than the individual error rates of $C_T$ and $C_I$, as failure of auditing requires both components to fail simultaneously to incorrectly satisfy Eq.~\ref{eq:condition}.
Modular ACs enable independent component training and validation prior to integration into the audit pipeline. 

\section{Experiments and Results}
\subsection{Data}
\label{subsec:data}
Experiments are performed using MIMIC-CXR \cite{mimic_paper}, containing chest X-rays (CXR), associated radiology reports, and disease labels for $\sim$228,000 studies of over 65,000 patients.
As common practice, we consider only the frontal-view images, and only the ``findings'' section of the reports, which describes evidence of pathologies and support devices. 
The labels 
include 14 classes (12 diseases, 1 support devices, 1 no findings) and
can be positive, uncertain, negative, or not mentioned. 
We report 5 out of 14 classes separately (Atelectasis, Cardiomegaly, Consolidation, Edema, Pleural Effusion) to facilitate comparison with literature \cite{chexpert_2019}. We divide data into splits for training (226,261), validating hyper-parameters (1,864), and testing performance (3,595). 
To train and evaluate the report generation, we further exclude studies that do not include a ``findings'' section, resulting in 155,322 training, 1,231 validation and 2,607 testing samples. Finally, to help address class imbalance and noisy labels, following past works \cite{Hou_ratchet_2021, chexpert_2019, miura_fact_comp_2021, tanida_interactive},
we consider uncertain disease labels as positive and class no-mention as negative.

\subsection{Evaluating the Report Generation Model}

\textbf{Experimental settings:} We here show that our report generation model, $GenX$, is of representative quality in comparison to previous work to serve as effective backbone for the study of our auditing framework. 
To build GenX we use ResNet-50 as the image encoder $f_I$, which we pre-train as a multi-label classifier of the 14 classes described above. Its penultimate layer of dimension $D_I=512$ is then projected to $M\!=\!10$ image token embeddings with $D_T\!=\!512$. These are concatenated with up to $L\!=\!512$ text embeddings, resulting from a GPT-2 tokeniser ($V\approx50,257$) \cite{radford_gpt2}, followed by a projection, $D_T\!=\!512$. The Transformer has $N_{layers}\!=\!8$, $N_{heads}\!=\! 8$, and dimension $D_{ff} \!=\! 2048$ of its feed-forward layer. GenX is then trained end-to-end (Fig.~\ref{fig:model}), learning parameters of the randomly initialised projections and Transformer, and fine-tuning the $f_I$ encoder.

\begin{table}[h]
\caption{Report generation performance of our model, GenX, compared to previous works. To assess reports' clinical correctness, we report macro and micro averages of F1 score, recall (R) and precision (P), averaged over the 5 most commonly reported classes (Atelect., Cardiom., Consolid., Edema, Pleural Eff. cf Sec.~\ref{subsec:data}) and over all 14 classes, in format (5 / 14 classes), to facilitate comparison with literature. 
Our model performs well in terms of F1 that quantifies clinical correctness, surpassed only by more complex frameworks using reinforcement \cite{Serra_2023} or additional labels \cite{miura_fact_comp_2021}, demonstrating that GenX is an effective backbone for our auditing study.
}
\centering
\renewcommand{\arraystretch}{1} 
\resizebox{\textwidth}{!}{
\begin{tabular}{
    @{}>{\raggedright}m{0.18\textwidth}
    *{6}{>{\centering\arraybackslash}m{0.14\textwidth}}
    *{3}{>{\centering\arraybackslash}m{0.0925\textwidth}}
    @{}
}
\toprule
&
\multicolumn{6}{c}{\textbf{Disease Classif. Metrics (Report)}}  &
\multicolumn{3}{c}{\multirow{2}{*}{\textbf{NLP Metrics}}} \\ 

\multirow{2}{*}{\textbf{Model}} &  \multicolumn{3}{c}{\textit{macro average}}& \multicolumn{3}{c}{\textit{micro average}} & \multicolumn{3}{c}{} \\ [-1ex]
\cmidrule(r){2-4} \cmidrule(r){5-7} \cmidrule(){8-10}

&  \textbf{F1(5/14)} & \textbf{R(5/14)} & \textbf{P(5/14)} & \textbf{F1(5/14)} & \textbf{R(5/14)} & \textbf{P(5/14)}  & \textbf{BL-1}  & \textbf{M} & \textbf{R-L}  \\

\midrule
R2Gen \cite{chen-2020}  & -/27.6 & -/27.3 & -/33.3  & -/- &-/-  &  -/- & 35.3 &  14.2 & 27.7 \\
Ratchet \cite{Hou_ratchet_2021}  & 38.8/- & -/- & -/-  & -/- &-/- & -/- & 23.2  & 10.1 & 24.0\\
\mbox{$\mathcal{M}^{2}$ Trans \cite{miura_fact_comp_2021}}  & 47.3/30.2 & 55.1/36.0 & 45.8/40.5  & 56.7/- & 65.1/- & 50.3/-   &-&-&-\\ 
RGRG \cite{tanida_interactive}   & -/- & -/- &-/-& 54.7/- & 61.7/- & 49.1/-  & 37.3 & 16.8 & 26.4 \\
\mbox{$A_{t}\texttt{+}$TE$\texttt{+}$RG\cite{Serra_2023}} & -/- & -/-  & -/- & -/53.7 & -/49.6 & -/58.5  & 49.0 &  21.3 & 40.6  \\
\textbf{GenX}  & 46.8/30.0 & 43.1/31.9 & 48.0/37.1 & 56.6/46.2 & 58.3/43.1 & 55.0/49.8 & 20.8 &  10.9 & 22.2 \\
\bottomrule
\end{tabular}}
\label{tbl:NLP_CE}
\end{table}

\textbf{Results:} 
Table~\ref{tbl:NLP_CE} reports test performance by our report generation model, along with previous works.
NLP metrics BLEU (BL)~\cite{bleu}, METEOR (M)~\cite{meteor} and \mbox{ROUGE-L} (R-L)~\cite{rouge_2004} assess generated language quality. 
More important for this study is the assessment of diagnosis-related semantics in generated reports, using disease-classification metrics such as F1 score, by comparing agreement of class labels, $C_T$, extracted from generated and reference reports via CheXbert \cite{chen-2020}.
Our report generator, GenX, achieves NLP metrics similar to  Ratchet \cite{Hou_ratchet_2021}, which has the most comparable architecture. Importantly, GenX achieves a high F1 score in comparison to Ratchet, R2Gen \cite{chen-2020}, even RGRG \cite{tanida_interactive} that needs additional labelled bounding boxes for training. It approaches F1 score by $\mathcal{M}^{2}$ Trans \cite{miura_fact_comp_2021} that trains using a more complex framework and reinforcement, and is surpassed by $A_{t}\texttt{+}$TE$\texttt{+}$RG\cite{Serra_2023} which uses extra labels for training.
We conclude that GenX produces reports of sufficient quality to serve as effective backbone for the following study of our auditing method. Two examples of generated reports are shown in Table \ref{tbl:good_egs}. The generated reports use realistic language, though they can occasionally miss a diagnosis. Thus a framework that can extract and audit report semantics is crucial.

\begin{table}[h]
     \caption{
     Examples of CXRs and their associated radiology report alongside the AI generated report from GenX. 
     }
     \centering
     \resizebox{\textwidth}{!}{
     \scriptsize
     
     \begin{tabular}
     {@{}>{\raggedright\arraybackslash}m{0.212\textwidth} m{0.45\textwidth}@{\hspace{0.02\textwidth}}
     m{0.45\textwidth}@{}}
     \toprule
    \centering\arraybackslash\textbf{X-ray} & \centering\arraybackslash\textbf{Reference Report} & \centering\arraybackslash\textbf{Generated Report} \\
    \cmidrule(r){1-1}\cmidrule(r){2-2}\cmidrule(){3-3}
     {\includegraphics[width=0.2\textwidth]{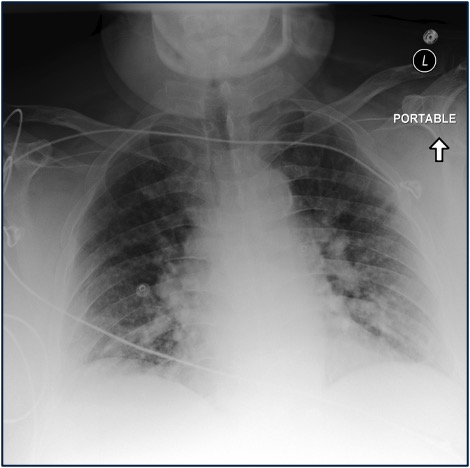}}
      & 
      The {\sethlcolor{cyan!20}\hl{lungs are low in volume}} but {\sethlcolor{cyan!20}\hl{without focal consolidation}}. Diffuse opacities likely reflect {\sethlcolor{cyan!20}\hl{mild pulmonary edema}}. There is {\sethlcolor{cyan!20}\hl{no pleural effusion or pneumothorax}}. The cardiac size and cardiac silhouette are obscured by low lung volumes. The mediastinal and hilar contours appear unremarkable.

      & 
      The {\sethlcolor{cyan!20}\hl{lung volumes are low}}. There is {\sethlcolor{cyan!20}\hl{mild pulmonary edema}}. There is {\sethlcolor{cyan!20}\hl{no focal consolidation, pleural effusion, or pneumothorax}}. The cardiomediastinal silhouette is normal.
      \\
      {\includegraphics[width=0.2\textwidth]{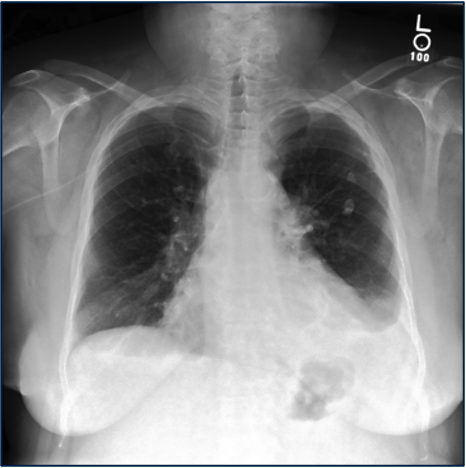}}&
      The ET and NG tubes have been removed. A right PICC line terminates in the low SVC. Calcified left lung nodules are unchanged. The lungs are otherwise clear except for {\sethlcolor{cyan!20}\hl{left basilar atelectasis}}. A {\sethlcolor{cyan!20}\hl{small left pleural effusion}} has developed. {\sethlcolor{cyan!20}\hl{Moderate cardiomegaly}} is unchanged.	&
      The {\sethlcolor{cyan!20}\hl{heart is mildly enlarged}}. There is mild unfolding of the thoracic aorta. There is a moderate-sized hiatal hernia. Streaky left basilar opacities suggest {\sethlcolor{cyan!20}\hl{minor atelectasis}}. Otherwise, the lungs appear clear. There is {\sethlcolor{orange!20}\hl{no definite pleural effusion}} or pneumothorax. Mild rightward convex curvature is centered along the mid thoracic spine. \\
      \bottomrule
      \end{tabular}}
      \label{tbl:good_egs}
\end{table}

\subsection{Evaluating the Auditing Framework for Generated Reports}

\textbf{Experimental settings:} 
We analyse whether including auxiliary auditing components, ACs, enable us to quality control the generated reports and filter out those that are more likely to contain errors.
We evaluate on the 5 diseases most commonly reported in the literature (Sec.~\ref{subsec:data}), from the CheXpert competition  \cite{chexpert_2019}. For this, we first train 5 independent disease-specific AC image classifiers ($AC_1$), where each is a ResNet-18. 
We then apply the framework as described in Sec.~\ref{subsec:method_audit} to the test-split. For each image, the 5 $AC_1$ models predict 5 disease labels $C_I^{(c)}$, one per disease $c$.
GenX generates a report for each test image and we extract 5 report-based class labels $C_T^{(c)}$ via Chexbert, for each disease.
For a given scan, per disease $c$, we compare image and report based labels, $C_I^{(c)}$ and $C_T^{(c)}$, to check if they fulfil Eq.~\ref{eq:condition}. We perform the check in two settings: First, without considering a confidence threshold for $p_{AC}$, satisfying only the first condition in Eq.~\ref{eq:condition}; Second, with a confidence threshold $t=0.8$ where Eq.~\ref{eq:condition} is fulfilled if $p_{AC}\!\geq\!0.8$, to investigate whether stricter auditing criteria results in more accurate reports
\
For each disease, the images that fulfil the above criteria are considered to have successfully passed the audit. For these images, we calculate average F1 score by comparing their report-based labels $C_T^{(c)}$ with the reference report labels, to determine whether the average score is higher than that of all unfiltered reports generated by GenX.
\
Finally, we repeat these experiments using a bigger ResNet-50 AC, trained as a multi-label image classifier of all 14 classes ($AC_{14}$). This is to evaluate whether it's beneficial to integrate a larger AC that learns from all 14 labels available per scan, or independent, disease-specific ACs. 

\begin{table}[h!]
\caption{Evaluation on test split of our auditing framework. For the 5 most commonly reported classes in literature, $GenX$ column reports F1 score of all generated reports ($C_T$ vs true label). $AC_1$ assesses image-based classification by 5 disease-specific ACs ($C_I$ vs true label). 
$GenX\!+\!AC_1^{t=0}$ and $GenX\!+\!AC^{t=0.8}_{1}$ show semantic factuality of generated reports that passed auditing for the specific disease against $AC_1$, satisfying Eq.~\ref{eq:condition} without the requirement for confidence $p_{AC}\!\geq\!t$ and when confidence over $t=0.8$ is required, respectively. 
Bold shows improvement by auditing over baseline GenX. 
Parentheses show percentage of reports that satisfy the auditing per disease. 
We also report results when auditing with a single multi-label AC trained for all 14 classes ($AC_{14}$ and $GenX\!+\!AC_{14}^{t=0}$). Developing independent ACs per disease is easier and more effective in practice.
}
\centering
\renewcommand{\arraystretch}{0.95}
\resizebox{\textwidth}{!}{
\begin{tabular}{
    @{}>{\raggedright}m{0.18\textwidth}
    *{1}{>{\centering\arraybackslash}m{0.09\textwidth}}
    *{1}{>{\centering\arraybackslash}m{0.11\textwidth}}
    *{2}{>{\centering\arraybackslash}m{0.21\textwidth}}
    *{1}{>{\centering\arraybackslash}m{0.11\textwidth}}
    *{1}{>{\centering\arraybackslash}m{0.21\textwidth}}
    @{}
}
\toprule
& \multicolumn{4}{c}{\textbf{F1 score (AC$_1$)}} & \multicolumn{2}{c}{\textbf{F1 score (AC$_{14}$)}} \\
\cmidrule(r){2-5} \cmidrule(l){6-7}
\textbf{Disease} & \textbf{AC$_{1}$} & \textbf{GenX} & \textbf{GenX+AC$^{t=0}_1$} & \textbf{GenX+AC$^{t=0.8}_1$} & \textbf{AC$_{14}$} & \textbf{GenX+AC$^{t=0}_{14}$} \\
\cmidrule(r){1-5}\cmidrule(l){6-7} 
Atelectasis & 55.6 & 42.2 & \textbf{52.6} (72) & \textbf{51.7} (30) & 37.1 & 31.8 (77) \\
Cardiomegaly & 57.6 & 66.8 & \textbf{73.7} (78) & \textbf{81.3} (34) & 40.3 & 53.3 (64) \\
Consolidation & 23.5 & 5.6 & \textbf{5.8} (76) & 1.6 (65) & 15.9 & 2.1 (89) \\
Edema & 62.1 & 55.6 & \textbf{64.6} (80) & \textbf{74.1} (45) & 53.1 & \textbf{60.2} (75) \\
\mbox{Pleural Effusion} & 71.9 & 63.8 & \textbf{74.7} (80) & \textbf{83.5} (57) & 61.8 & \textbf{64.1} (82) \\
\cmidrule(r){1-5}\cmidrule(l){6-7} 
\textit{Average} & 54.1 & 46.8 & \textbf{54.3} & \underline{\textbf{58.4}} & 41.6 & 42.3 \\
\bottomrule
\end{tabular}}
\label{tbl:rep_aud}
\end{table}

\textbf{Results:} Table~\ref{tbl:rep_aud} shows the results. Class labels $C_T$ extracted from reports generated by GenX achieve an average F1 score of $46.8$ over the 5 diseases. Reports that agree with $AC_1$, fulfilling the auditing without the extra requirement for confidence $p_{AC\!}\geq\!t$, achieve average F1 score $54.3$. These benefits are realised over a significant percentage of the reports (72-80\% across the diseases), while the rest would be deferred for user inspection and error correction in an actual workflow.
Reports that pass the stricter auditing and 
additionally fulfil the confidence requirement $p_{AC}\!\geq\!0.8$ present even less errors, with average F1 score $58.4$. This restricts the number of reports that pass the audit (30-65\%), in return for higher reliability.
In comparison to the unfiltered reports of GenX, the reports that fulfil the audit show increased F1-scores for 4 out of 5 studied diseases. The exception is Consolidation, the class where both GenX and AC image classifiers have very low performance. Interestingly, for 3 out of 5 diseases, the reports that successfully pass the audit show F1-scores that are even higher than the auditing $AC_1$ classifiers.
These results show that measuring consistency between semantics of reports and auditing classifiers is a promising framework.

Finally, we find that a single, multi-label classifier, trained for all 14-classes as $AC_{14}$, achieves significantly lower F1 scores for image classification, which also leads to lower effectiveness of the auditing. Although $AC_{14}$ is a ResNet-50, larger than ResNet-18s used for the 5 $AC_1$ classifiers,
it is difficult to develop it at the level of disease-specific classifiers. This demonstrates the benefits of modularity, with independently developed AC for each concept of interest.


\section{Conclusion}
We introduced a framework to quality control AI-generated radiology reports with respect to semantics of interest via auxiliary auditing components. 
Our study focused on the task of report generation from chest-Xrays, detecting potential semantic errors in diagnosis-related report content, using auxiliary disease classifiers and assessing consistency between labels inferred from image and report.
Experiments show that reports fulfilling the auditing criteria exhibit fewer errors, with average F1 score that can even exceed the score of the auditing components. The framework is generic and future work could explore it for auditing semantic concepts beyond disease classification, such as for description of pathology location or volume, using regression models as ACs for such properties.

The limitation of the framework, as with any filtering pipeline for quality-control, is the inherent trade-off between reducing the number of reports that successfully pass the audit and can be used for down-stream workflow, in exchange for their higher reliability. Future work could improve performance in both aspects by integrating more potent models for inferring image-based and report-based labels, thereby ensuring greater consistency and reliability from their comparison.

\section{Acknowledgements}
Hermione Warr and Yasin Ibrahim are supported by the EPSRC Centre for Doctoral Training in Health Data Science (EP/S02428X/1).

%
%
%
\bibliographystyle{splncs04}
\bibliography{bib}

\end{document}